\documentclass[10pt,twocolumn,numbers]{article}

\usepackage{graphicx}
\usepackage{amssymb}
\usepackage{amsthm,amsmath,dsfont,bbm}
\usepackage{mathtools}

\usepackage{lineno}

\usepackage[colorlinks=true, citecolor=blue]{hyperref}
\usepackage[linesnumbered,lined,ruled,commentsnumbered]{algorithm2e}
\usepackage{algcompatible}

\usepackage{wrapfig}
\usepackage[numbers,sort&compress]{natbib}
\usepackage{comment}
\usepackage{longtable}
\usepackage{multirow}
\usepackage{bbding}
  
\usepackage[bottom]{footmisc}
\usepackage{booktabs}
\usepackage{pifont}
\usepackage{float}
\usepackage{subfigure}
\usepackage{wrapfig}
\usepackage{framed,xcolor}
\usepackage{newfloat}
\usepackage[xcolor,framemethod=TikZ]{mdframed}
\usepackage{amsthm}
\usepackage{thmtools}
\usepackage{authblk}
\patchcmd{\thebibliography}{\section*{\refname}}{}{}{}
\usepackage{tikz}
\usetikzlibrary{arrows,shapes,chains,calc,patterns,decorations.pathreplacing}
\usepackage{stmaryrd}
\usepackage{fancyhdr}
\usepackage{subfigure}
\usepackage{comment}
\def\BibTeX{{\rm B\kern-.05em{\sc i\kern-.025em b}\kern-.08em
		T\kern-.1667em\lower.7ex\hbox{E}\kern-.125emX}}

\renewcommand{\headrulewidth}{2pt}

\newlength\FHoffset

\fancyheadoffset{\FHoffset}

\newlength\FHleft
\newlength\FHright

\newbox\FHline
\setbox\FHline=\hbox{\hsize=\paperwidth%
	\hspace*{\FHleft}%
	\rule{\dimexpr\headwidth-\FHleft-\FHright\relax}{\headrulewidth}\hspace*{\FHright}%
}

\newtheoremstyle{theoremdd}
{\topsep}
{\topsep}
{\itshape}
{0pt}
{\fontfamily{cmss}\selectfont\bfseries}
{.}
{ }
{\thmname{#1}\thmnumber{ #2}\thmnote{ (#3)}}

\theoremstyle{theoremdd}

\mdfdefinestyle{myStyle}{roundcorner=5pt,backgroundcolor=brown!20,linecolor=red!40!black,linewidth=1.5pt}

\usepackage{titlesec}
\titleformat*{\section}{\fontfamily{cmss}\selectfont\large\bfseries\color{red!40!black}}
\titleformat*{\subsection}{\fontfamily{cmss}\selectfont\normalsize\bfseries\color{red!40!black}}
\titleformat*{\subsubsection}{\fontfamily{cmss}\selectfont\normalsize\color{red!40!black}}
\usepackage[left=0.69 in, right=0.69 in, top=1.1 in, bottom=1.3 in]{geometry}
\graphicspath{{./Figures/}}

\newcommand\blfootnote[1]{%
	\begingroup
	\renewcommand\thefootnote{}\footnote{#1}%
	\addtocounter{footnote}{-1}%
	\endgroup
}
\renewcommand\abstractname{\fontfamily{cmss}\selectfont\normalsize\bfseries\color{red!40!black}\textbf{Abstract}}

\renewenvironment{abstract}{%
	\centering\small
	\list{}{\leftmargin1.5cm \rightmargin\leftmargin}
	\item\relax
	
	\begin{mdframed}[]
		\item[\hskip\labelsep\scshape\abstractname.]%
	}{%
	\end{mdframed}
	\endlist \par\bigskip
}
\makeatletter
\patchcmd{\@maketitle}{\LARGE \@title}{\fontfamily{cmss}\selectfont\LARGE\color{red!40!black}\@title}{}{}
\makeatother

\graphicspath{{./Figures/}}

\begin{document}

		
		
		\title{Learning Traffic Speed Dynamics from Visualizations}
		
			

\author[1,2]{Bilal Thonnam Thodi}
\author[1,2]{Zaid Saeed Khan$^\dagger$}
\author[1,2]{Saif Eddin Jabari}
\author[1,2]{M\'onica Men\'endez}

\affil[1]{New York University Tandon School of Engineering, Brooklyn NY, U.S.A.}
\affil[2]{New York University Abu Dhabi, Saadiyat Island, P.O. Box 129188, Abu Dhabi, U.A.E.}

\date{}


\twocolumn[
\begin{@twocolumnfalse}
	
\maketitle	

\begin{abstract}
	Space-time visualizations of macroscopic or microscopic traffic variables is a qualitative tool used by traffic engineers to understand and analyze different aspects of road traffic dynamics. We present a deep learning method to learn the macroscopic traffic speed dynamics from these space-time visualizations, and demonstrate its application in the framework of traffic state estimation. Compared to existing estimation approaches, our approach allows a finer estimation resolution, eliminates the dependence on the initial conditions, and is agnostic to external factors such as traffic demand, road inhomogeneities and driving behaviors. Our model respects causality in traffic dynamics, which improves the robustness of estimation. We present the high-resolution traffic speed fields estimated for several freeway sections using the data obtained from the Next Generation Simulation Program (NGSIM) and German Highway (HighD) datasets. We further demonstrate the quality and utility of the estimation by inferring vehicle trajectories from the estimated speed fields, and discuss the benefits of deep neural network models in approximating the traffic dynamics.
	
	\medskip
	
	\textbf{\fontfamily{cmss}\selectfont\color{red!40!black} Keywords}: Traffic flow dynamics, traffic anisotropy, deep learning.
\end{abstract}
\bigskip
\end{@twocolumnfalse}
]

		
	
	
	

\section{Motivation and Overview}
\label{sec:introduction}
 \blfootnote{$^\dagger$ Corresponding author. Email: \url{zaid.khan@nyu.edu}.}

Quantifying the road traffic conditions is an essential prerequisite for several Intelligent Transport System (ITS) applications such as urban traffic signal control, ramp-metering, variable speed limits, and connected and autonomous vehicles \cite{Kaidi2016TState,yang2017perimeter,li2019position,bilal2021bsp}. Due to limited real-time traffic observations from the field, this is often performed using data assimilation with macroscopic or microscopic traffic flow models, statistical inference from a large amount of historical traffic data, or, more recently, structured machine learning techniques \cite{seo2017traffic}.

In this study, we demonstrate the learning of traffic dynamics from visualizations that are easily interpretable by traffic engineers. A typical example is the macroscopic traffic speed field defined over the space-time plane, such as the one shown in Fig.~\ref{fig:inp-out}(b). This plot contains a wealth of traffic information $-$ the traffic speed, the congested part(s) of the road section, and the velocity and extent of backward propagating stop-and-go waves. Although these visualizations can be theoretically reproduced to some extent using microscopic traffic flow models, those have a high computational cost, need lots of data, and require the problem to be well-posed. The key contribution of this paper is to show that these visualizations can be used to reverse engineer and learn the underlying traffic dynamics.

We use deep learning tools to learn the speed fields from visualizations, partly motivated by their success in image analytics and computer vision applications \cite{krizhevsky2012imagenet,LeCun2015dl}. Here, we define the macroscopic speed field over a discrete space-time plane of fine mesh size; we use $10$m $\times$ $1$s without loss of generality. For our learning process, we show the neural network model samples of the speed field containing only partial information, and train it to reconstruct the complete macroscopic speed field. In practice, this partial information can be provided by moving probe vehicles or stationary traffic sensors. Here, we assume the partial information is obtained from the probe vehicles and has a limited coverage, e.g. $5\%$. The speed field samples required for training the neural network model are generated using traffic simulation.

The learning model used in this study is a Deep Convolutional Neural Network (Deep CNN) stacked into an encoder-decoder architecture. Deep CNN models have properties favorable for learning the space-time correlations between traffic speeds \cite{benkraouda2020traffic}. The architecture of the Deep CNN model is further modified to explicitly capture the directions of traffic causality over the space-time plane. In the experiments, we show high-resolution traffic speed fields estimated for several real freeway segments and other traffic variables that can be inferred from these estimated speed maps. This general learning framework can address a broad class of related traffic problems $-$ state estimation, data imputation and prediction; we limit our focus to traffic state estimation.

\section{Related Works} \label{sec:review}

Existing approaches for traffic estimation are broadly classified into $1)$ model-based data assimilation techniques, $2)$ data-driven statistical methods, and $3)$ structured learning methods. The first approach combines the predictions of a traffic flow model with real-time traffic observations using an exogenous filter, and is the most popular in practice \cite{nantes2016real,jabari2013gauss,hoogen2012lagrang,wang2005kf}. These techniques honor basic traffic physics and are more robust to outliers. However, these methods are restricted by the representational power of the selected traffic flow model, and require additional inputs such as initial and boundary conditions and traffic flow model parameters. Furthermore, their estimations are often coarse grained, i.e., most studies assume an estimation interval of $100-500$ m $\times$ $5-15$ minutes (with an exception of a few recent studies \cite{li2020nonlinear}) and are not suitable for arbitrary space-time resolutions. The latter is a drawback of the numerical scheme used in solving the traffic flow model; for e.g., in the Eulerian coordinate system, the mesh sizes in the space and time dimensions are restricted by the Courant-Fredich-Lewy (CFL) condition \cite{treiber2013traffic}. 

Data-driven methods use large amounts of historical traffic data to synthesize patterns and use them for future predictions. These methods don't utilize any traffic domain knowledge. The popular choices for learning algorithms include Deep Neural Networks \cite{benkraouda2020traffic,polson2017deep,wang2016lstm}, Support Vector Machines \cite{wang2013short} and ARIMA \cite{Kumar2015sarima}. They have been shown to perform better than model-based approaches in capturing the daily and weekly traffic flow pattern, but their practicalilty is limited by their large data requirements, lack of robustness and poor interpretability. More recently, structured machine learning techniques have been proposed to improve the data-driven methods by incorporating traffic domain knowledge into the learning framework \cite{jabari2020sparse,jabari2019learning,kaidi2019queueest,liu2020learning,zhang2020hybrid,huang2020physics}. In deep learning, this is often done by adding a regularizer. For e.g., \cite{huang2020physics,liu2020learning} used the LWR macroscopic flow model, and \cite{zhang2020hybrid} used the integral form of the vehicle conservation law as the cost function regularizer. The reported benefits include feasible and robust estimation along with lower data requirements. However, only macroscopic traffic flow models are amenable to such integration, which again limit their representational capacity.

The visualization-based learning method outlined in this paper overcomes some of the limitations mentioned above. First, our approach allows the speed field to be quantified with a fine mesh size and an arbitrary space-time resolution. This is beneficial if one requires speed fields for smaller road sections but at longer time intervals or vice-versa. Second, the estimation method does not assume knowledge of initial or boundary conditions; it is able to produce an estimate for any given input speed field at low computational cost (approximately the cost associated with a single forward pass of the Deep CNN model). Third, learning the dynamics over interpretable visualizations allow to easily incorporate domain knowledge to the learning model, e.g., causality (the anisotropy of traffic flow). Finally, using simulated data as a surrogate to real data is an implicit way of incorporating complex traffic behaviors (obtained from the traffic flow models used in the simulation) into the learning model.

\section{Learning Model} \label{sec:model}

\subsection{Problem Setting}

\begin{figure}[!hbt]
\centering
\includegraphics[width=0.4\textwidth]{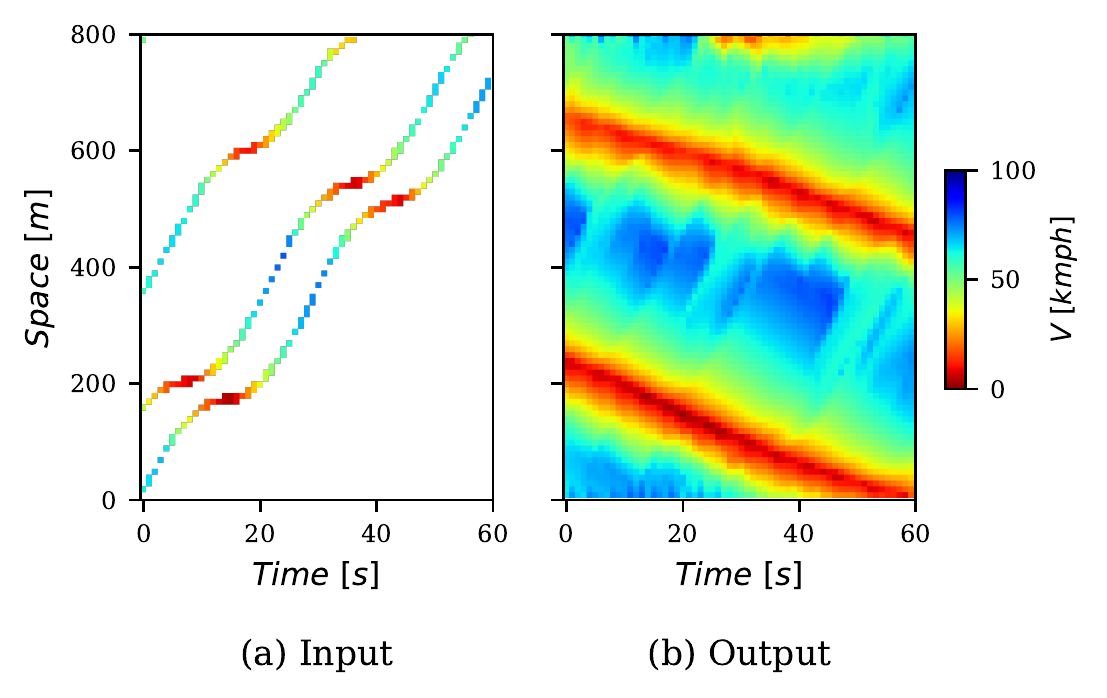}
\caption{Input speed field $\mathbf{v}^{\rm p}$ and output speed field $\mathbf{v}^{\rm f}$.}
\label{fig:inp-out}
\end{figure}

The macroscopic speed field is defined over a space-time plane of dimension $|X| \times |T|$, where $X$ and $T$ are the sets of discretizations in space and time coordinates, respectively. The complete macroscopic speed field is denoted by $\mathbf{v}^{\rm f} \in [0, V_{\rm max}]^{|X|\times|T|}$ where $V_{\rm max}$ is the maximum traffic speed; this is the output of the learning model. The input to the model is the partially filled speed field $\mathbf{v}^{\rm p}$. A sample pair of input and output speed fields for a road section is shown in Fig.~\ref{fig:inp-out}.

An immediate challenge here is the representation of input speed field $\mathbf{v}^{\rm p}$. Since the traffic speed is a scalar variable, a matrix representation of the input speed field $\mathbf{v}^{\rm p}$ cannot distinguish between cells with zero traffic speed (i.e. congested cells) and cells with missing information about the traffic speed. A more natural way to represent the input speed field is in the form of a three dimensional RGB (Red-Green-Blue) array as opposed to an array of single dimension speed values. For e.g., with reference to Fig.~\ref{fig:inp-out}, the RGB representation of a missing cell is $(0,0,0)$ and that of congested cell is $(255,0,0)$. 
Thus, the input speed field is $\mathbf{v}^{\rm p} \in [0,1]^{|X| \times |T| \times 3}$, where the RGB values are normalized between $0$ and $1$. 

The learning problem aims to approximate a non-linear operator $\mathcal{G}$ that maps the input speed field to the output speed field, i.e., $\mathcal{G}: \mathbf{v}^{\rm p} \rightarrow \mathbf{v}^{\rm f}$.

\subsection{Deep Convolutional Neural Network (CNN) Model}

\begin{figure}[!hbt]
\centering
\includegraphics[width=0.47\textwidth]{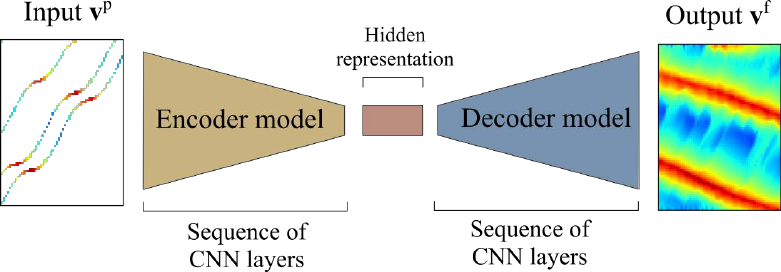}
\caption{Deep Convolutional Neural Network (CNN) model.}
\label{fig:cnn_mod}
\end{figure}

The mapping operator $\mathcal{G}$ is approximated using an encoder-decoder neural network architecture as shown in Fig.~\ref{fig:cnn_mod}. The encoder and decoder models are each composed of a sequence of Convolutional Neural Network (CNN) layers. The encoder model takes in the input speed field $\mathbf{v}^{\rm p}$  and produces a higher dimensional hidden representation, which is then decoded by the decoder model to form the output speed field $\mathbf{v}^{\rm f}$. As depicted in Fig.~\ref{fig:cnn_mod}, the hidden representation extracts the essential spatio-temporal traffic information from the sparse input speed field $\mathbf{v}^{\rm p}$.

The Deep CNN model has specific properties that make it suitable for learning traffic dynamics. Traffic speed at a point $(x,t)$ in the space-time plane only depends on the traffic conditions in its immediate neighborhood. \emph{Sparse connectivity} of the CNN model ensures this by having local neuron connections, with local being defined as the CNN kernel size. The parameter sharing property of the CNN model implies that the learned traffic features can occur anywhere over the space-time plane, i.e, they are space-time invariant. As a result, the Deep CNN model is independent of $|X|$ and $|T|$, and can be applied to any road section length and time period.

\subsection{Anisotropic Deep CNN Model}

\begin{figure}[!hbt]
\centering
\includegraphics[width=0.47\textwidth]{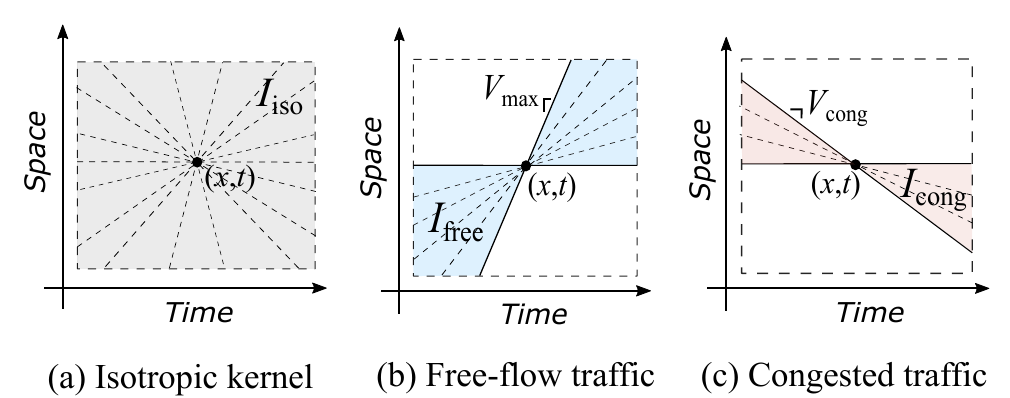}
\caption{Anisotropic CNN kernel capturing traffic causality. $V_{\rm max}$ is the maximum traffic speed and $V_{\rm cong}$ is the backward shockwave speed.}
\label{fig:aniso}
\end{figure}

We further modify the architecture of the Deep CNN model to explicitly capture the asymmetric direction of causality in traffic. Convolutional operators in CNNs treat information coming from all directions equally. This is depicted in Fig.~\ref{fig:aniso}(a), where the traffic speed at $(x,t)$ has a space-time correlation defined by $I_{\rm iso}$. This implies that a variation in the traffic speed at $(x,t)$ can propagate at infinite velocity.

In traffic, variations at $(x,t)$ only propagate at finite velocities. Moreover, these propagation velocities depend on the traffic state at $(x,t)$, i.e. whether it is in free-flow or is congested \cite{hoogendorn2013ani,trieber2011filter,treiber2002filter}. The former property is called hyperbolicity and the latter is called anisotropy. In free-flow traffic, speed variations propagate in the direction of traffic flow whereas in the congested traffic, speed variations propagate in the opposite direction of traffic flow. The anisotropic property is specific to traffic flow, and is due to drivers reacting more to leading vehicles than to following vehicles. The resulting space-time correlations are $I_{\rm free}$ and $I_{\rm cong}$, as depicted in Figs.~\ref{fig:aniso}(b) and (c). We modify the isotropic kernel of the Deep CNN model to reflect this.

\subsection{Simulated Data and Training}

The data required for training the anisotropic Deep CNN model is generated using the Vissim traffic microsimulator \cite{vissim}. We simulate a real multi-lane freeway segment with three levels of traffic demand inputs $-$ free-flow, slow-moving and congested traffic. The simulation model is calibrated to replicate the corresponding real driving conditions. The simulation is run for $2$ hours for each scenario, and vehicle trajectory data is recorded. The macroscopic traffic speed field is then calculated from the vehicle trajectory data. The input-output training sample pairs are then derived from this speed field; we have a total of $20,000+$ training samples.

We use the TensorFlow framework \cite{tensorflow} to train our anisotropic Deep CNN. The encoder and decoder models each consist of three CNN layers. After calibration, the kernel configuration of the model is: $(5\times 5\times 40), (7\times 7\times 48), (7\times 7\times 32), (5\times 5\times 48), (5\times 5\times 40), (9\times 9\times 56)$ and $(7 \times 7 \times 1)$. We train the model using stochastic gradient optimization with a mean squared error loss function. Note that the training procedure is carried out offline.

\section{Results and Discussion} \label{sec:results}

\subsection{Speed Field Reconstruction}

\begin{figure}[!hbt]
\centering
\includegraphics[width=0.4\textwidth]{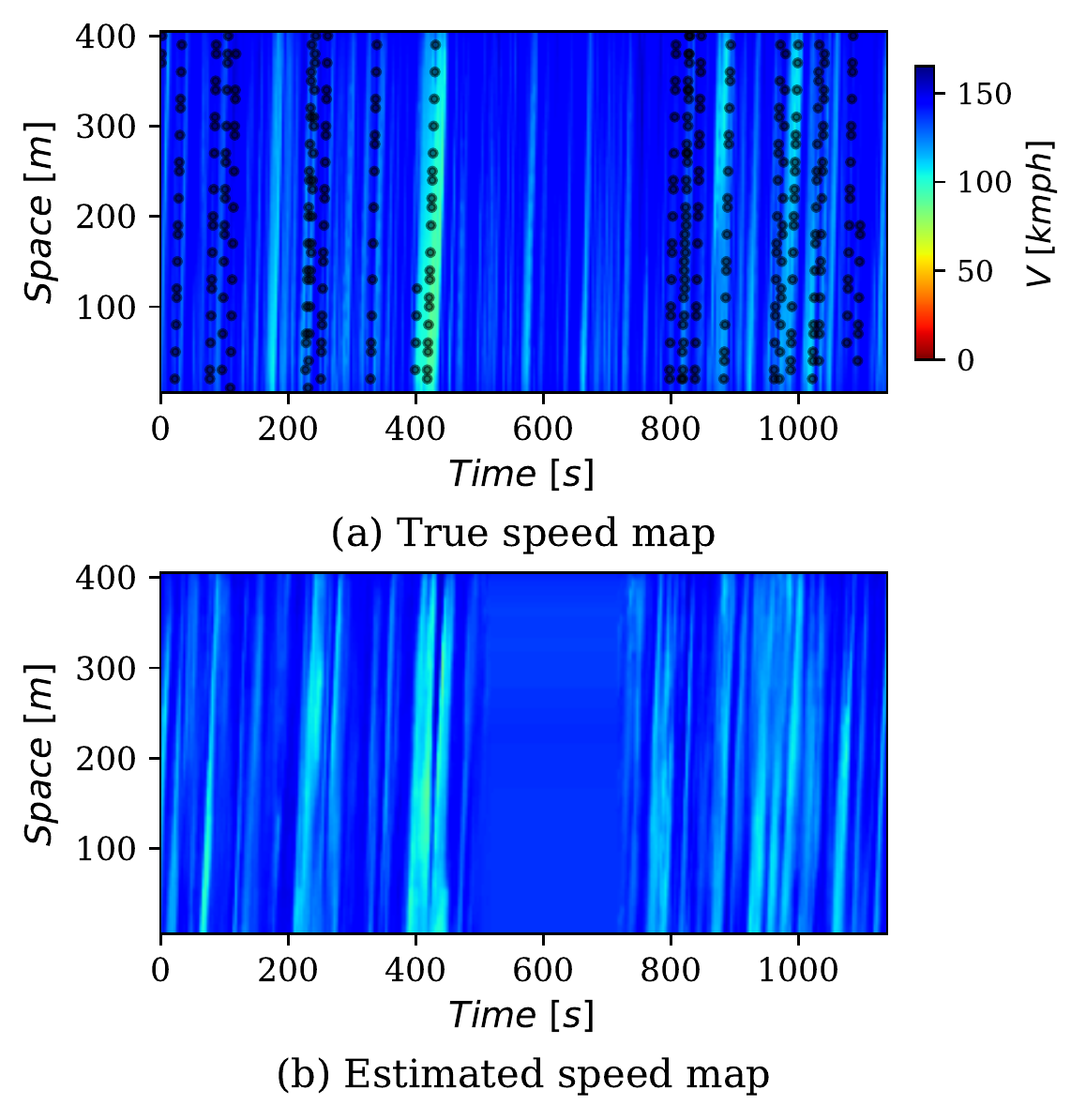}
\caption{Speed field estimation in free-flow traffic (Highway 44 lane 6 of the HighD dataset). The RMSE is $14.60$ kmph.}
\label{fig:recon1}
\end{figure}

\begin{figure}[!hbt]
\centering
\includegraphics[width=0.4\textwidth]{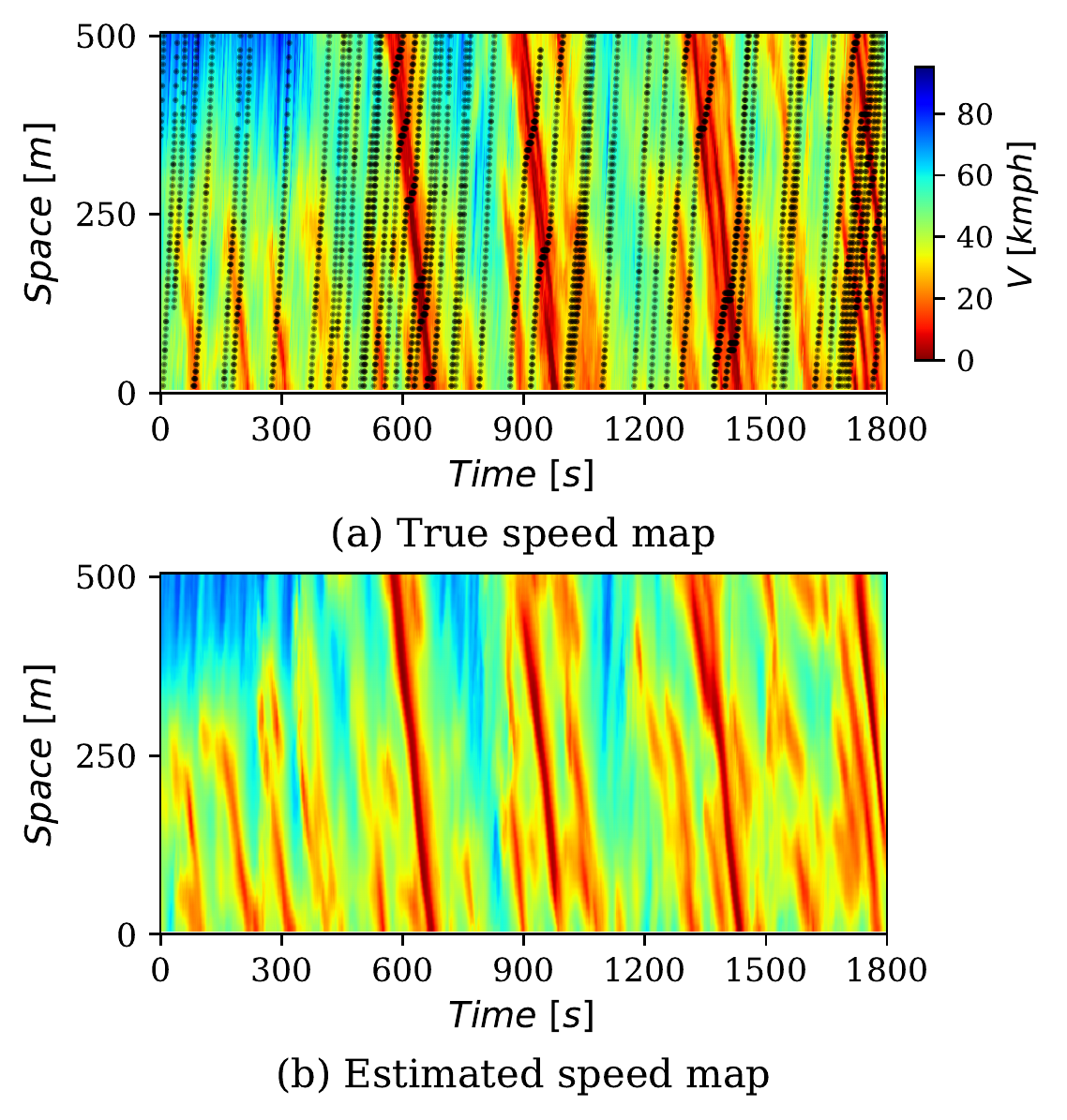}
\caption{Speed field estimation in slow-moving traffic with breakdowns (Highway 25 lane 4 of the HighD dataset). The RMSE is $8.5$ kmph.}
\label{fig:recon3}
\end{figure}

\begin{figure}[!thb]
\centering
\includegraphics[width=0.4\textwidth]{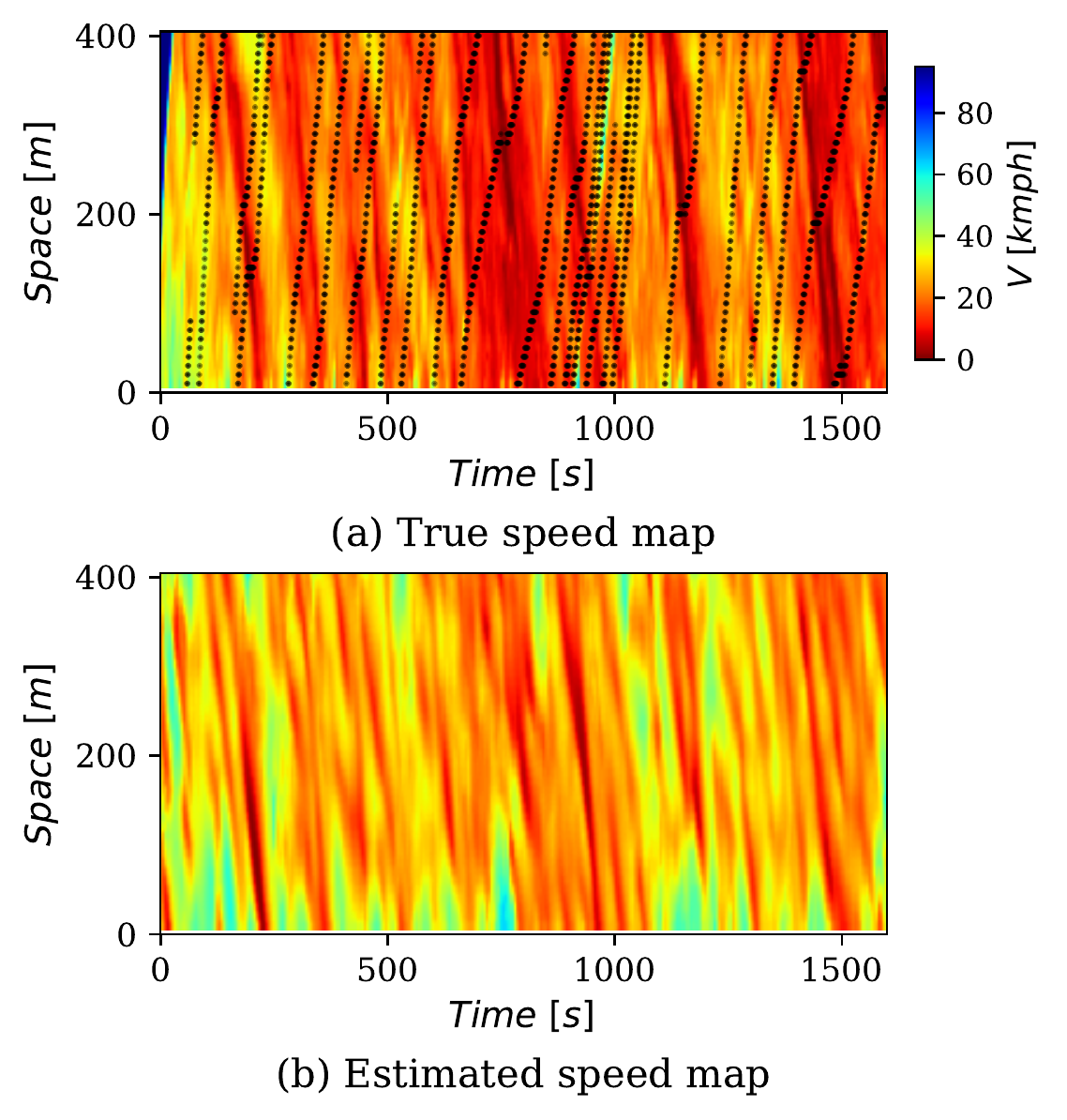}
\caption{Speed field estimation in heavily congested traffic (US-101 highway lane 3 of the NGSIM dataset). The RMSE is $12.7$ kmph.}
\label{fig:recon5}
\end{figure}

We evaluate the performance of the anisotropic Deep CNN model using real freeway data obtained from the Next Generation Simulation program (NGSIM) dataset \cite{ngsim} and the German Highway (HighD) dataset \cite{krajewski2018highd}. A few sample estimated speed fields under different traffic conditions are shown in Figs.~\ref{fig:recon1} through \ref{fig:recon5}. For all estimations, the mesh size is $10$m $\times$ $1$s, and the probe coverage is  $5\%$. The input probe vehicle trajectories are shown as black dotted curves. 

A preliminary observation of the reconstructed speed fields shows that they are indeed similar to the true speed fields despite the limited coverage of the probe vehicles. The traffic dynamics they produce look feasible by observation, and the existence of different traffic states in different parts of the time-space plane is captured with reasonable accuracy.

Fig.~\ref{fig:recon1} shows that free-flow traffic is estimated well by the Deep CNN. The turquoise bands representing relatively slower free-flow traffic are reconstructed accurately when there are probe vehicles present. However, given the low rate of probe coverage among the already sparse vehicles present in free-flow traffic, there are large areas of the space-time plane without coverage. In these areas, the model interpolates from the closest probe trajectories while assuming that there are no slowdowns in the uncovered area. This is a reasonable conclusion in the absence of any information, but it leads to the estimated speed being higher than the real speed when there are unobserved slower moving vehicles present (see the area around $600$s). The Root Mean Squared Error (RMSE) of the estimation is quite high at $14.60$ kmph, but the relative RMSE is only $8.80\%$, which implies that the estimation is accurate and acceptable.

Fig.~\ref{fig:recon3} shows an example of slow-moving traffic with backward propagating waves. The existence and extent of these waves is closely replicated by the model. For example, the shockwaves in the first $300$s of the true speed field dissipate around $250$m, whereas later shockwaves do not dissipate. Both of these phenomena can also be seen in the estimated speed field. In slow-moving traffic, vehicles interact with each other more frequently, and the collective dynamics can be inferred effectively from the trajectories of a few vehicles. Coupled with the effect of having better coverage due to the increased density of vehicles, this leads to the RMSE of slow-moving traffic being low (around $7-9$ kmph).

Fig.~\ref{fig:recon5} shows an example where the model does not reconstruct the dynamics very accurately. The true speed field shows heavy congestion, with vehicles moving at very low speeds for long periods. The estimated speed field does show the congestion, but with more prominent waves and slightly higher speeds. This is because the speed distribution in this real traffic sample is very different from the speed distribution in our simulated training set; the model has not seen such consistently-slow-moving traffic while training. Whenever a probe trajectory crosses a slightly faster moving region, the model extrapolates the speed field in the vicinity to be similarly faster, whereas it is not usually the case. This can be seen at several points in time near the beginning of the road segment ($0-50$m). However, even under this unfamiliar scenario, the reconstruction is reasonable, and the RMSE is not very high. Note that this is an extreme case shown for illustration; in typical examples of congested traffic, where the dynamics are not this drastically different from those observed in training, the model performs better (RMSE around $8-10$ kmph).

\subsection{Trajectory Inference}

\begin{figure}[!thb]
\centering
\includegraphics[width=2.5in]{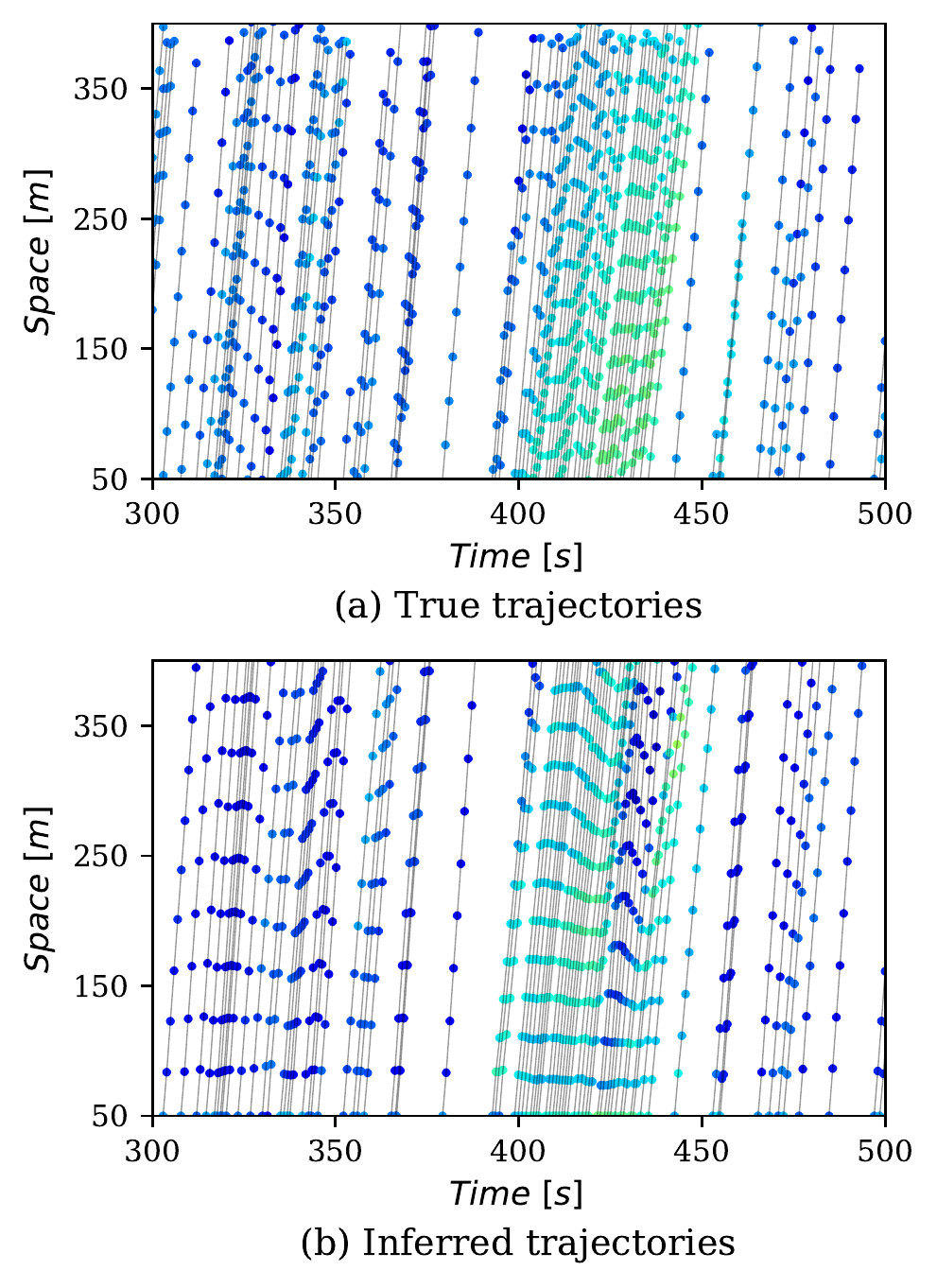}
\caption{Vehicle trajectories inferred from part of the estimated speed field of free-flow traffic in Fig.~\ref{fig:recon1}.}
\label{fig:traj1}
\end{figure}

\begin{figure}[!thb]
\centering
\includegraphics[width=2.5in]{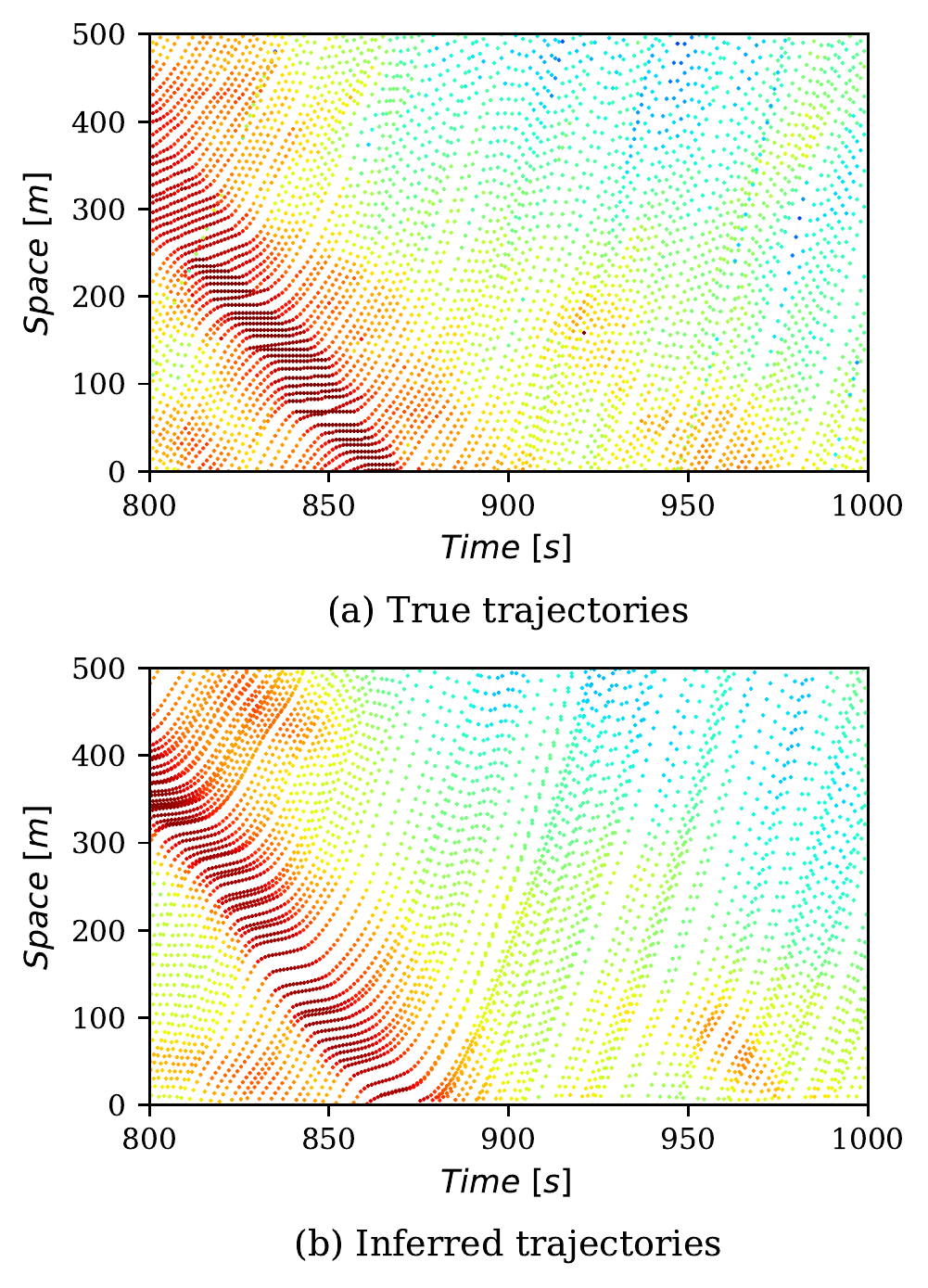}
\caption{Vehicle trajectories inferred from part of the estimated speed field of congested traffic in Fig.~\ref{fig:recon3}.}
\label{fig:traj2}
\end{figure}

The estimated speed field can also be used to infer the trajectories of all the vehicles. This is done by taking the entry time of each vehicle (which is an additional input that can be determined from a stationary traffic sensor) and then tracing its position through the speed field. Figs.~\ref{fig:traj1} and \ref{fig:traj2} show the true and inferred trajectories corresponding to a free-flow and congested scenario respectively; there is good agreement in both cases. 
Furthermore, the inferred trajectories do not overlap, implying that the estimation does not produce physically infeasible traffic patterns. The inferred trajectories are smooth versions of the actual trajectories (compare the stop-and-go behavior in Figs.~\ref{fig:traj2}(a) and (b)). These observations demonstrate that the anisotropic Deep CNN model learned to produce traffic speed fields that honor basic traffic physics without explicitly imposing any physical constraints. One can further use these inferred trajectories to estimate other traffic variables such as density and headways.

\subsection{Hidden Representation}

\begin{figure}[!thb]
\centering
\includegraphics[width=2.3in]{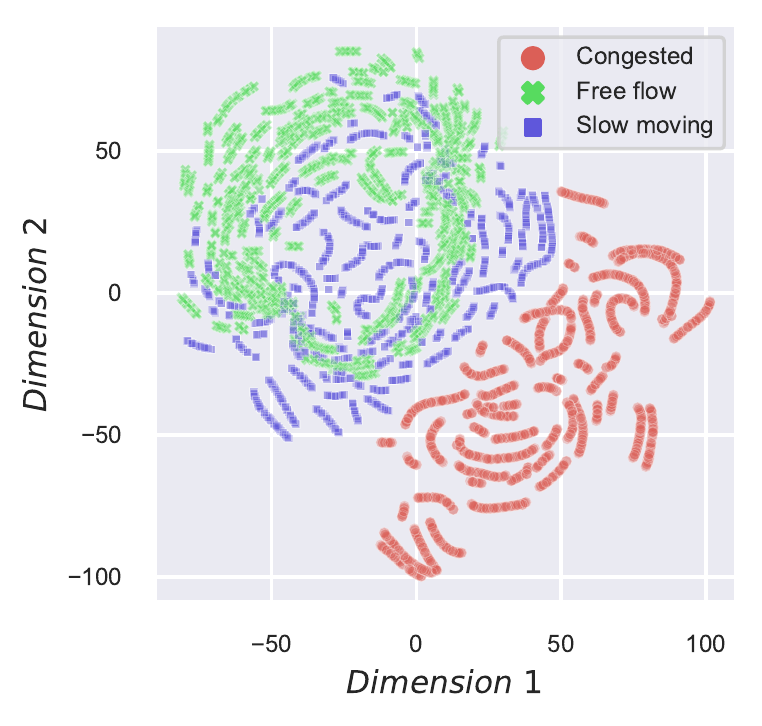}
\caption{Hidden representation of $3000$ input speed field samples (output of the encoder model) projected to two dimensions and color coded with their respective traffic states.}
\label{fig:latent}
\end{figure}

In order to demonstrate that our model has learned to distinguish between different traffic states, we visualize the information it extracts from its input. We project the hidden representation of $3000$ unseen input speed field samples (i.e., the output from the encoder part of the anisotropic Deep CNN) onto two dimensions using the t-SNE algorithm, a popular non-linear dimensionality reduction technique \cite{hinton2008tsne}. This is shown in Fig.~\ref{fig:latent}, where each scatter point corresponds to a single input speed field. The two-dimensional scatter points are then color coded with their respective traffic state $-$ free-flow, slow-moving or congested.

We notice that the congested traffic samples form an isolated cluster, while there is some overlap among the free-flow and slow-moving traffic samples. These clusters become more distinct with higher probe coverage. The presence of these well-defined clusters in the hidden representation shows that the deep neural network model can infer the traffic state from the available input observations and produces traffic waves accordingly; it does not require any additional inputs. In other words, a neural network model can solve the under-determined estimation problem $-$ a major benefit compared to other machine learning tools. The other benefits highlighted in this study, such as eliminating the dependence on initial or boundary conditions, the high resolution of the estimation, and the sparse observation setting, are also partly enabled by use of a deep neural network model as the function approximator.

\section{Conclusion} \label{sec:concl}

In this study, we present a deep learning method to estimate road traffic dynamics from space-time diagrams of macroscopic traffic speed fields. The space-time visualizations of traffic variables such as speed or density reflect different features of traffic conditions, such as stop-and-go waves, congested parts of the road, and the extent of queued vehicles. Our objective is to exploit these interpretable visualizations to learn the underlying traffic dynamics while being agnostic to external factors such as traffic demand, road geometry, and weather conditions. We demonstrate this within the framework of traffic state estimation and highlight the key benefits it can achieve over traditional estimation techniques.

Our speed estimation model is a Deep Convolutional Neural Network (Deep CNN) that uses only sparse probe vehicle trajectory observations to reconstruct the speed field over a given discretized space-time plane with fine resolution. The CNN uses an anisotropic kernel which is designed to incorporate traffic causality by capturing information from only the relevant directions in free-flow and congested traffic. We train the model using simulated traffic data under various demand conditions, in order to implicitly incorporate the dynamics of the traffic flow model used in the simulation. We test the model using real data from multiple datasets with varied traffic characteristics. Our results show that the anisotropic Deep CNN performs well in terms of estimation error, given the sparsity of the probe coverage and the sometimes unfamiliar characteristics of real traffic. In particular, the reconstructed traffic dynamics are plausible and do not violate basic traffic physics principles. We further show how to use the estimated speed field to infer the trajectories of all vehicles.

Learning the dynamics from physically meaningful visualization has significant benefits for traffic engineering applications. It allows us to easily integrate heterogeneous traffic data sources as input to the model. For instance, dynamic features such as weather condition, road visibility, and temporary road inhomogeneities (lane closures or variable speed limits) can be visualized over a space-time diagram and be used as inputs to the learning model. This application can also be extended to interrupted traffic facilities like urban arterial, where we can represent the signal timing as a horizontal speed band in the space-time diagram. Furthermore, the computational requirement for producing a high-resolution estimate is relatively low when working in the image space compared to the raw representation of speeds and positions of all vehicles on the road. Lastly, the use of deep learning tools, which are universal function approximators, can overcome the curse of dimensionality problems arising in traffic engineering applications.

\section*{Acknowledgment}
This work was supported by the NYUAD Center for Interacting Urban Networks (CITIES), funded by Tamkeen under the NYUAD Research Institute Award CG001 and by the Swiss Re Institute under the Quantum Cities\textsuperscript{TM} initiative. The views expressed in this article are those of the authors and do not reflect the opinions of CITIES or its funding agencies.

\appendix
\gdef\thesection{Appendix \Alph{section}}

	
	
	
\bibliographystyle{plainnat}
\bibliography{tse-itsc2021}

\begin{thebibliography}{32}
\providecommand{\natexlab}[1]{#1}
\providecommand{\url}[1]{\texttt{#1}}
\expandafter\ifx\csname urlstyle\endcsname\relax
  \providecommand{\doi}[1]{doi: #1}\else
  \providecommand{\doi}{doi: \begingroup \urlstyle{rm}\Url}\fi

\bibitem[Abadi et~al.(2016)Abadi, Barham, Chen, Chen, Davis, Dean, Devin,
  Ghemawat, Irving, Isard, et~al.]{tensorflow}
Mart{\'\i}n Abadi, Paul Barham, Jianmin Chen, Zhifeng Chen, Andy Davis, Jeffrey
  Dean, Matthieu Devin, Sanjay Ghemawat, Geoffrey Irving, Michael Isard, et~al.
\newblock Tensorflow: A system for large-scale machine learning.
\newblock In \emph{12th $\{$USENIX$\}$ symposium on operating systems design
  and implementation ($\{$OSDI$\}$ 16)}, pages 265--283, 2016.

\bibitem[Benkraouda et~al.(2020)Benkraouda, Thodi, Yeo, Menendez, and
  Jabari]{benkraouda2020traffic}
Ouafa Benkraouda, Bilal~Thonnam Thodi, Hwasoo Yeo, Monica Menendez, and
  Saif~Eddin Jabari.
\newblock Traffic data imputation using deep convolutional neural networks.
\newblock \emph{IEEE Access}, 8:\penalty0 104740--104752, 2020.

\bibitem[Fellendorf(1994)]{vissim}
M~Fellendorf.
\newblock Vissim: Ein instrument zur beurteilung verkehrsabh{\"a}ngiger
  steuerungen (vissim: a tool to analyse vehicle actuated signal control).
\newblock \emph{the German Research Foundation for transport and roads (FGSV),
  Kassel}, pages 31--38, 1994.

\bibitem[Huang and Agarwal(2020)]{huang2020physics}
Jiheng Huang and Shaurya Agarwal.
\newblock Physics informed deep learning for traffic state estimation.
\newblock In \emph{IEEE Intelligent Transportation Systems Conference}, 2020.

\bibitem[Jabari and Liu(2013)]{jabari2013gauss}
S.E. Jabari and H.~Liu.
\newblock A stochastic model of traffic flow: {G}aussian approximation and
  estimation.
\newblock \emph{Transportation Research Part B: Methodological}, 47:\penalty0
  15--41, 2013.

\bibitem[Jabari et~al.(2019)Jabari, Dilip, Lin, and
  Thonnam~Thodi]{jabari2019learning}
S.E. Jabari, D.~Dilip, D.~Lin, and B.~Thonnam~Thodi.
\newblock Learning traffic flow dynamics using random fields.
\newblock \emph{IEEE Access}, 7:\penalty0 130566--130577, 2019.

\bibitem[Jabari et~al.(2020)Jabari, Freris, and Dilip]{jabari2020sparse}
S.E. Jabari, N.~Freris, and D.~Dilip.
\newblock Sparse travel time estimation from streaming data.
\newblock \emph{Transportation Science}, 54\penalty0 (1):\penalty0 1--20, 2020.

\bibitem[Krajewski et~al.(2018)Krajewski, Bock, Kloeker, and
  Eckstein]{krajewski2018highd}
Robert Krajewski, Julian Bock, Laurent Kloeker, and Lutz Eckstein.
\newblock The highd dataset: A drone dataset of naturalistic vehicle
  trajectories on german highways for validation of highly automated driving
  systems.
\newblock In \emph{2018 21st International Conference on Intelligent
  Transportation Systems (ITSC)}, pages 2118--2125. IEEE, 2018.

\bibitem[Krizhevsky et~al.(2012)Krizhevsky, Sutskever, and
  Hinton]{krizhevsky2012imagenet}
A.~Krizhevsky, I.~Sutskever, and G.~Hinton.
\newblock Imagenet classification with deep convolutional neural networks.
\newblock In \emph{Advances in Neural Information Processing Systems}, pages
  1097--1105, 2012.

\bibitem[Kumar and Vanajakshi(2015)]{Kumar2015sarima}
S.~Vasantha Kumar and Lelitha Vanajakshi.
\newblock Short-term traffic flow prediction using seasonal {ARIMA} model with
  limited input data.
\newblock \emph{European Transport Research Review}, 7\penalty0 (3), June 2015.
\newblock \doi{10.1007/s12544-015-0170-8}.
\newblock URL \url{https://doi.org/10.1007/s12544-015-0170-8}.

\bibitem[LeCun et~al.(2015)LeCun, Bengio, and Hinton]{LeCun2015dl}
Yann LeCun, Yoshua Bengio, and Geoffrey Hinton.
\newblock Deep learning.
\newblock \emph{Nature}, 521\penalty0 (7553):\penalty0 436--444, May 2015.
\newblock \doi{10.1038/nature14539}.
\newblock URL \url{https://doi.org/10.1038/nature14539}.

\bibitem[Li and Jabari(2019)]{li2019position}
L.~Li and S.E. Jabari.
\newblock Position weighted backpressure intersection control for urban
  networks.
\newblock \emph{Transportation Research Part B: Methodological}, 128:\penalty0
  435--461, 2019.

\bibitem[Li et~al.(2020)Li, Yang, and Jabari]{li2020nonlinear}
Wenqing Li, Chuhan Yang, and Saif~Eddin Jabari.
\newblock Nonlinear traffic prediction as a matrix completion problem with
  ensemble learning.
\newblock \emph{arXiv preprint arXiv:2001.02492}, 2020.

\bibitem[Liu et~al.(2020)Liu, Barreau, and Johansson]{liu2020learning}
John Liu, Matthieu Barreau, and Karl~H Johansson.
\newblock Learning-based traffic state reconstruction using probe vehicles.
\newblock \emph{arXiv preprint arXiv:2011.05031}, 2020.

\bibitem[Nantes et~al.(2016)Nantes, Ngoduy, Bhaskar, Miska, and
  Chung]{nantes2016real}
Alfredo Nantes, Dong Ngoduy, Ashish Bhaskar, Marc Miska, and Edward Chung.
\newblock Real-time traffic state estimation in urban corridors from
  heterogeneous data.
\newblock \emph{Transportation Research Part C: Emerging Technologies},
  66:\penalty0 99--118, 2016.

\bibitem[Polson and Sokolov(2017)]{polson2017deep}
N.~Polson and V.~Sokolov.
\newblock Deep learning for short-term traffic flow prediction.
\newblock \emph{Transportation Research Part C: Emerging Technologies},
  79:\penalty0 1--17, 2017.

\bibitem[Seo et~al.(2017)Seo, Bayen, Kusakabe, and Asakura]{seo2017traffic}
T.~Seo, A.~Bayen, T.~Kusakabe, and Y.~Asakura.
\newblock Traffic state estimation on highway: {A} comprehensive survey.
\newblock \emph{Annual Reviews in Control}, 43:\penalty0 128--151, 2017.

\bibitem[Thodi et~al.(2021)Thodi, Chilukuri, and Vanajakshi]{bilal2021bsp}
Bilal~Thonnam Thodi, Bhargava~Rama Chilukuri, and Lelitha Vanajakshi.
\newblock An analytical approach to real-time bus signal priority system for
  isolated intersections.
\newblock \emph{Journal of Intelligent Transportation Systems}, 0\penalty0
  (0):\penalty0 1--23, 2021.
\newblock \doi{10.1080/15472450.2020.1797504}.
\newblock URL \url{https://doi.org/10.1080/15472450.2020.1797504}.

\bibitem[Treiber and Helbing(2002)]{treiber2002filter}
M.~Treiber and D.~Helbing.
\newblock Reconstructing the spatio-temporal traffic dynamics from stationary
  detector data.
\newblock \emph{Cooper@tive Tr@nsport@tion Dyn@mics}, 1\penalty0 (3):\penalty0
  3--1, 2002.

\bibitem[Treiber and Kesting(2013)]{treiber2013traffic}
M.~Treiber and A.~Kesting.
\newblock \emph{Traffic Flow Dynamics: {D}ata, Models, and Simulation}.
\newblock Springer-Verlag, Berlin, 2013.

\bibitem[Treiber et~al.(2011)Treiber, Kesting, and Wilson]{trieber2011filter}
Martin Treiber, Arne Kesting, and R.~Eddie Wilson.
\newblock Reconstructing the traffic state by fusion of heterogeneous data.
\newblock \emph{Computer-Aided Civil and Infrastructure Engineering},
  26\penalty0 (6):\penalty0 408--419, 2011.

\bibitem[{United States Department of Transportation}(2006)]{ngsim}
{United States Department of Transportation}.
\newblock {NGSIM—Next Generation Simulation}, 2006.
\newblock URL \url{https://ops.fhwa.dot.gov/trafficanalysistools/ngsim.htm}.

\bibitem[van~der Maaten and Hinton(2008)]{hinton2008tsne}
Laurens van~der Maaten and Geoffrey Hinton.
\newblock Visualizing data using t-sne.
\newblock \emph{Journal of Machine Learning Research}, 9\penalty0
  (86):\penalty0 2579--2605, 2008.
\newblock URL \url{http://jmlr.org/papers/v9/vandermaaten08a.html}.

\bibitem[van Wageningen-Kessels et~al.(2013)van Wageningen-Kessels, van't Hof,
  Hoogendoorn, van Lint, and Vuik]{hoogendorn2013ani}
Femke van Wageningen-Kessels, Bas van't Hof, Serge~P. Hoogendoorn, Hans van
  Lint, and Kees Vuik.
\newblock Anisotropy in generic multi-class traffic flow models.
\newblock \emph{Transportmetrica A: Transport Science}, 9\penalty0
  (5):\penalty0 451--472, 2013.
\newblock \doi{10.1080/18128602.2011.596289}.
\newblock URL \url{https://doi.org/10.1080/18128602.2011.596289}.

\bibitem[Wang et~al.(2016)Wang, Gu, Wu, Liu, and Xiong]{wang2016lstm}
J.~Wang, Q.~Gu, J.~Wu, G.~Liu, and Z.~Xiong.
\newblock Traffic speed prediction and congestion source exploration: {A} deep
  learning method.
\newblock In \emph{2016 IEEE 16th International Conference on Data Mining
  (ICDM)}, pages 499--508, 2016.

\bibitem[Wang and Shi(2013)]{wang2013short}
Jin Wang and Qixin Shi.
\newblock Short-term traffic speed forecasting hybrid model based on
  chaos--wavelet analysis-support vector machine theory.
\newblock \emph{Transportation Research Part C: Emerging Technologies},
  27:\penalty0 219--232, 2013.

\bibitem[Wang and Papageorgiou(2005)]{wang2005kf}
Yibing Wang and Markos Papageorgiou.
\newblock Real-time freeway traffic state estimation based on extended kalman
  filter: a general approach.
\newblock \emph{Transportation Research Part B: Methodological}, 39\penalty0
  (2):\penalty0 141 -- 167, 2005.
\newblock ISSN 0191-2615.
\newblock \doi{https://doi.org/10.1016/j.trb.2004.03.003}.
\newblock URL
  \url{http://www.sciencedirect.com/science/article/pii/S0191261504000438}.

\bibitem[Yang and Menendez(2019)]{kaidi2019queueest}
K.~Yang and M.~Menendez.
\newblock Queue estimation in a connected vehicle environment: {A} convex
  approach.
\newblock \emph{IEEE Transactions on Intelligent Transportation Systems},
  20\penalty0 (7):\penalty0 2480--2496, 2019.

\bibitem[Yang et~al.(2016)Yang, Guler, and Menendez]{Kaidi2016TState}
K.~Yang, S.~Guler, and M.~Menendez.
\newblock Isolated intersection control for various levels of vehicle
  technology: {C}onventional, connected, and automated vehicles.
\newblock \emph{Transportation Research Part C: Emerging Technologies},
  72:\penalty0 109--129, 2016.

\bibitem[Yang et~al.(2017)Yang, Zheng, and Menendez]{yang2017perimeter}
Kaidi Yang, Nan Zheng, and Monica Menendez.
\newblock Multi-scale perimeter control approach in a connected-vehicle
  environment.
\newblock \emph{Transportation Research Procedia}, 23:\penalty0 101 -- 120,
  2017.
\newblock ISSN 2352-1465.
\newblock \doi{https://doi.org/10.1016/j.trpro.2017.05.007}.

\bibitem[{Yuan} et~al.(2012){Yuan}, {van Lint}, {Wilson}, {van
  Wageningen-Kessels}, and {Hoogendoorn}]{hoogen2012lagrang}
Y.~{Yuan}, J.~W.~C. {van Lint}, R.~E. {Wilson}, F.~{van Wageningen-Kessels},
  and S.~P. {Hoogendoorn}.
\newblock Real-time lagrangian traffic state estimator for freeways.
\newblock \emph{IEEE Transactions on Intelligent Transportation Systems},
  13\penalty0 (1):\penalty0 59--70, 2012.
\newblock \doi{10.1109/TITS.2011.2178837}.

\bibitem[Zhang et~al.(2020)Zhang, Yuan, and Yang]{zhang2020hybrid}
Zhao Zhang, Yun Yuan, and Xianfeng Yang.
\newblock A hybrid machine learning approach for freeway traffic speed
  estimation.
\newblock \emph{Transportation research record}, 2674\penalty0 (10):\penalty0
  68--78, 2020.

\end{thebibliography}
	
	
	
	
	
	

\end{document}